\documentclass[letterpaper]{article} 
\usepackage{aaai23}  
\usepackage{times}  
\usepackage{helvet}  
\usepackage{courier}  
\usepackage[hyphens]{url}  
\usepackage{graphicx} 
\usepackage{natbib}  
\usepackage{caption} 
\usepackage{multirow}
\usepackage{subfigure}
\usepackage{amsfonts}
\usepackage{algorithm}
\usepackage{algorithmic}

\usepackage{newfloat}
\usepackage{listings}
\DeclareCaptionStyle{ruled}{labelfont=normalfont,labelsep=colon,strut=off} 
\floatstyle{ruled}
\newfloat{listing}{tb}{lst}{}
\floatname{listing}{Listing}
\nocopyright

\title{Distribution Aligned Feature Clustering for Zero-Shot Sketch-Based Image Retrieval}
\author{
    Yuchen Wu, Kun Song, Fangzheng Zhao, Jiansheng Chen, Huimin Ma
     \thanks{
     Corresponding author, mhmpub@ustb.edu.cn \newline \indent \indent
     Code is available at https://github.com/curiosity654/ClusterRetr.
     }
}
\affiliations{
    University of Science and Technology Beijing
}

\usepackage{bibentry}

\begin{document}

\maketitle

\begin{abstract}

Zero-Shot Sketch-Based Image Retrieval (ZS-SBIR) is a challenging cross-modal retrieval task. In prior arts, the retrieval is conducted by sorting the distance between the query sketch and each image in the gallery. However, the domain gap and the zero-shot setting make neural networks hard to generalize. This paper tackles the challenges from a new perspective: utilizing gallery image features. We propose a Cluster-then-Retrieve (ClusterRetri) method that performs clustering on the gallery images and uses the cluster centroids as proxies for retrieval. Furthermore, a distribution alignment loss is proposed to align the image and sketch features with a common Gaussian distribution, reducing the domain gap. Despite its simplicity, our proposed method outperforms the state-of-the-art methods by a large margin on popular datasets, e.g., up to $31\%$ and $39\%$ relative improvement of mAP@all on the Sketchy and TU-Berlin datasets.
\end{abstract}
\section{Introduction}
Sketch-based image retrieval (SBIR) is an application-driven task aiming to retrieve similar images based on hand-drawn sketches. With the development of deep learning and large-scale labeled datasets, SBIR can achieve satisfying performance. However, with the exponential growth of image content online, the classes to be retrieved have outnumbered any existing SBIR dataset, while the evaluation methodology does not guarantee generalization on novel classes. To this end, zero-shot SBIR (ZS-SBIR)~\cite{c:81,c:80}  is proposed, which aims to retrieve categories unseen at the training stage.

The essence of zero-shot sketch-based image retrieval is a generalizable mapping from sketch to image. The primary challenges of this task are twofold: huge domain gap and large intra-class variance. Early works~\cite{c:82,c:80,c:83} employ generative models or leverage semantic information to alleviate the domain gap. SAKE~\cite{c:86} first uses the ImageNet ~\cite{c:89} pretrained teacher model to preserve useful features learned on real-world images, which significantly boosts retrieval performance. But the two challenges are not taken into account. TVT~\cite{c:130} first models the global structural information in SBIR using Vision Transformer and perform inter- and intra-modal alignment using multi-modal hypersphere learning.


In the above methods, retrieval is conducted by sorting the distance between the query image and each image in the gallery. The query image is a sketch with highly abstract strokes and no color feature. In contrast, target objects in natural images appear at various angles and are surrounded by complicated backgrounds. Especially in the zero-shot setting, the model can only find images with similar geometry information, significantly harming the performance. Thanks to the ImageNet pretrained model, the gallery feature vectors generated by SBIR algorithms have highly informative local structures. Natural clusters of classes can be found in the feature space, providing valuable information for this task.

In this paper, we propose a Cluster-then-Retrieve method (ClusterRetri) that uses the structure of gallery feature space to aid the retrieval process. We use K-Means clustering algorithm to divide the gallery feature into clusters and use the centers as proxies for retrieval. Pre-clustering can effectively reduce the intra-class variance. With color and texture features, objects of different views can be grouped and retrieved. Moreover, the proposed subspace clustering can reduce the noise introduced by outliers in clusters, further improving the performance. In addition, we adopt a distribution alignment loss that minimizes the KL divergence between feature vectors and random features sampled from a Gaussian distribution. This loss can align the image and sketch features with a common Gaussian distribution, reducing the domain gap between images and sketches.

Our contributions can be summarized as follows:
\begin{itemize}
    \item This is the first study that leverages gallery information to mitigate intra-class variance in ZS-SBIR.
    \item We propose a simple yet effective pipeline to utilize gallery information. Specifically, the pre-clustering, subspace division, and feature fusion can discover groups of images and help alleviate the domain gap.
    \item We adopt Kullback--Leibler divergence that aligns the feature distribution of sketches and images with a prior distribution and further improves ZS-SBIR performance.
    \item Extensive experiments conducted on two popular datasets demonstrate the effectiveness of our method. The proposed methods can provide up to 31\% improvement on the baseline and outperform the state-of-the-art methods by a large margin.
\end{itemize}

\section{Related Work}

\subsection{SBIR and ZS-SBIR}
The fundamental problem of the SBIR task is to project the hand-drawn sketches and the real-world images into a common metric space, where samples of the same category are close to each other. The earliest works extract hand-crafted features from the sketches and match them with the edge maps of the natural images~\cite{c:90,c:91,c:92,c:93,c:94}. In recent years, with the prevalence of deep neural networks (DNNs), different architectures of neural networks are introduced into this field~\cite{c:95,c:96,c:97,c:98,c:99,c:100} and achieve superb results. However, the closed-set setting of SBIR does not meet the demand of large-scale applications. To this end, zero-shot setting of SBIR~\cite{c:80,c:81} is proposed. Following zero-shot learning approaches, ZS-SBIR studies~\cite{c:81,c:101} use semantic information to bridge the seen and unseen classes. Other methods employ generative models, such as generative adversarial networks (GANs)~\cite{c:83} and variational auto-encoders (VAEs)~\cite{c:80} to learn a mapping from sketches to images. However, the inherent sparsity of sketch and intra-class variance of images make it difficult for semantic information to transfer, or for the network to learn a generalizable mapping on the unseen classes. Following works~\cite{c:85,c:84} start to use metric losses to improve retrieval performance, but the pair-wise losses often require large batch size or complex sample mining techniques to perform well. Different from the above approaches, our Cluster-then-Retrieve method reduces intra-class variance by utilizing gallery information.

\subsection{Clustering in Information Retrieval}

Data clustering has been adopted for information retrieval for a long time.~\cite{c:117} propose a hierarchic clustering method that significantly improves the efficiency of information retrieval. The study also proposes the cluster hypothesis: \textit{the associations between documents convey information about the relevance of documents to requests}, and pointed out clustering's potential of improving retrieval effectiveness. Based on this statement, many studies~\cite{c:118,c:119,c:120} try to investigate whether clustering can bring performance improvement to retrieval algorithms. However, although improvements could be achieved through manually selecting clusters~\cite{c:122}, clustering-based retrieval fails to gain better results in practice~\cite{c:121}. In contrast, clustering has been widely applied to feature quantization~\cite{c:123,c:124,c:125} , enhancing retrieval efficiency on large-scale datasets. In the scenario of ZS-SBIR, estimating distances in a single domain could be more accurate than that between different domains, so the potential of cluster-based retrieval can be revealed.

\section{Method}

In this section, we first recap the problem formulation of ZS-SBIR, and then we examine our motivation, which is to leverage gallery information to alleviate the large intra-class variance among gallery images. We then illustrate our Cluster-then-Retrieve pipeline that uses cluster centroids as proxies for retrieval and distribution alignment loss that alleviate the domain gap between sketches and images.

\subsection{Problem Formulation}

Consider a dataset consisting of the training set and the testing set. We denote the training set as $\mathcal{D}^{tr}=\{\mathcal{I}^{tr},\mathcal{S}^{tr}\}$, where $\mathcal{I}^{tr}$ and $\mathcal{S}^{tr}$ are subsets of images and sketches. Similarly, the testing set is defined as $\mathcal{D}^{te}=\{\mathcal{I}^{te},\mathcal{S}^{te}\}$.

Let $\mathcal{I}^{tr}=\{(\mathcal{I}_i,y_i)\mid y_i \in \mathcal{C}^{tr}\}^{N_1}_{i=1}$ and $\mathcal{S}^{tr}=\{(\mathcal{S}_i,y_i)\mid y_i \in \mathcal{C}^{tr}\}^{N_2}_{i=1}$, where $y_i$ is the class label and $\mathcal{C}^{tr}$ is the set of training classes. The images and labels are used to train the model to learn discriminative features. After training, the image encoder generates feature vectors of the gallery images and query sketches, preparing for the retrieval. The testing images and sketches are $\mathcal{I}^{te}=\{(\mathcal{I}_i,y_i)\mid y_i \in \mathcal{C}^{te}\}^{N_3}_{i=1}$, $\mathcal{S}^{te}=\{(\mathcal{S}_i,y_i)\mid y_i \in \mathcal{C}^{te}\}^{N_4}_{i=1}$, and the corresponding feature vectors are denoted as $\mathcal{F}^\mathcal{I}=\{\mathbf{x}_i\}^{N_3}_{i=1}$, $\mathcal{F}^\mathcal{S}=\{\mathbf{q}_i\}^{N_4}_{i=1}$. During testing, given a sketch feature $\mathbf{q}_i$, its distances with gallery features $\mathcal{F}^\mathcal{I}$ are calculated and sorted, giving an optimal ranking of gallery images. In the zero-shot setting, the testing classes and the training classes do not overlap, i.e., $\mathcal{C}^{tr} \cap \mathcal{C}^{te}=\emptyset$.

\subsection{Motivation: Leveraging Gallery Information in ZS-SBIR}
\begin{figure*}
    \centering
    \includegraphics[width=0.85\textwidth]{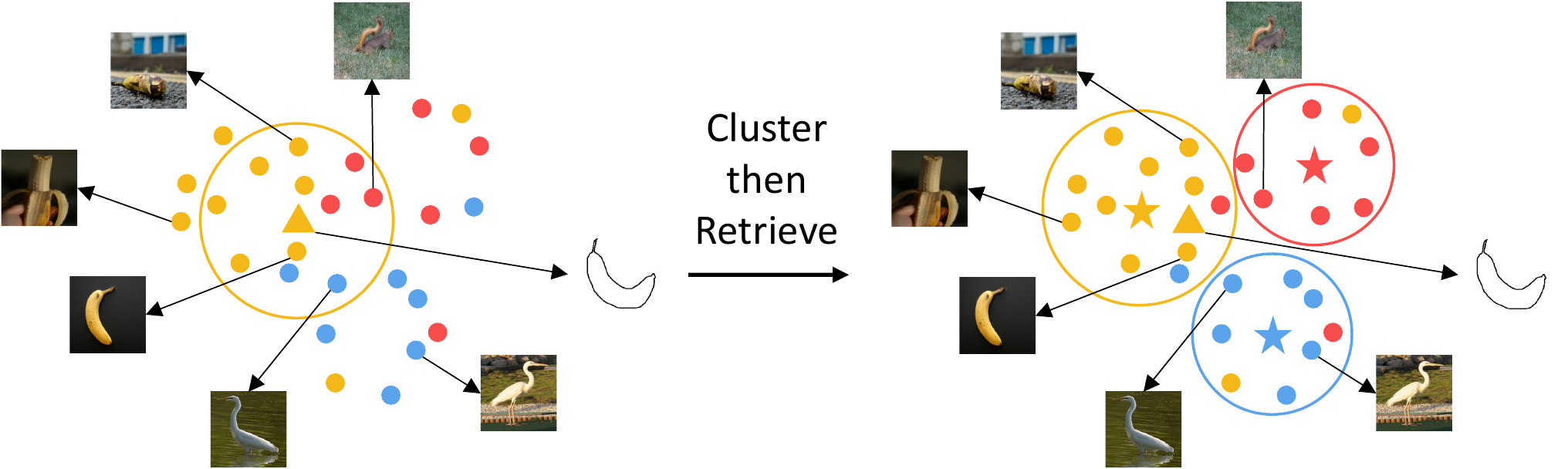}
    \caption{An illustration of our motivation. The sample query is an unpeeled banana, while the bananas in the gallery appear in various angles and forms. There are also samples of other classes having outlines similar to the banana, acting as distractors. Clustering before retrieval can substantially reduce the intra-class variance and number of distractors, improving retrieval results.}
    \label{fig:motivation}
\end{figure*}

The ZS-SBIR is inherently a difficult problem because of its form: a cross-domain zero-shot metric learning task. We calculate and sort the distance between sketch and gallery images for each query. The highly abstract strokes provide merely sparse geometry information, so the model can only infer from the image edges and sketch strokes. However, as Fig.\ref{fig:motivation} shows, objects in gallery images appear in various forms and are surrounded by complex backgrounds. Without the help of large pretrained models or side information, this problem seems intractable. 

Thanks to the ImageNet pretrained model, we can obtain an informative feature space on the gallery images, where objects of different views can be grouped. Therefore, we perform clustering on the gallery and use the cluster centroids as proxies for retrieval. 


\subsection{Cluster-then-Retrieve Method}
\subsubsection{Clustering and Retrieval}
\begin{figure*}[t]
    \centering
    \includegraphics[width=0.9\textwidth]{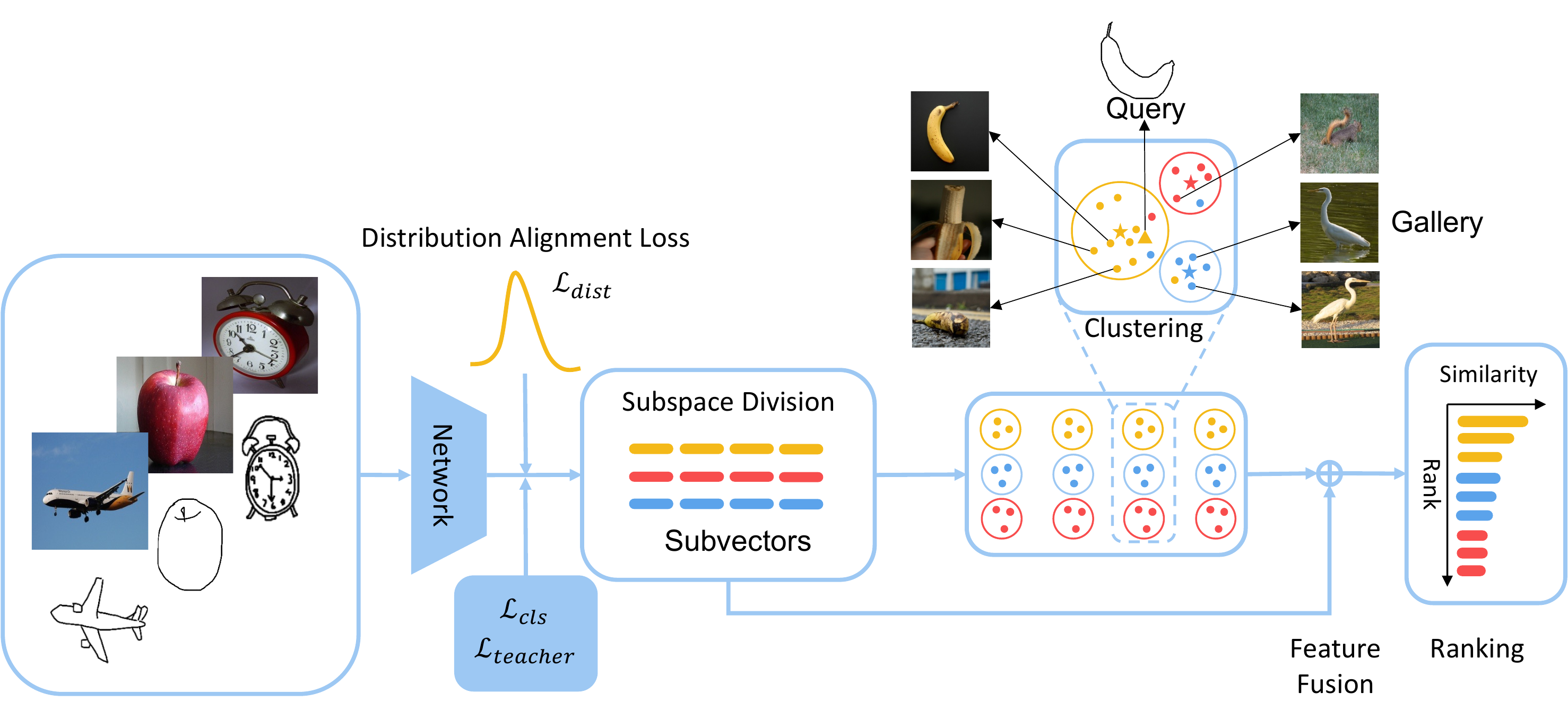}
    \caption{An overview of our pipeline. CSE-ResNet-50 projects the images and sketches into a common feature space. Classification and distillation losses guide the model to learn discriminative features. KL divergence is used to align the feature distribution and reduce domain gap. The output features are randomly divided into subvectors, and clustering is performed in each subspace. The reproduction vectors and original features are fused using a convex combination. Finally, gallery images are sorted according to the distances between fused and query features to obtain the retrieval rank.}
    \label{fig:my_label}
\end{figure*}

Consider a gallery of images $\mathcal{I}$, and their features $\mathcal{F}^\mathcal{I}=\{\mathbf{x}^\mathcal{I}_i\in \mathbb{R}^D\}^N$, where $\mathbf{x}^\mathcal{I}_i$ is the feature vector of each image in the gallery, $D$ denotes the dimension of features, and $N$ denotes the number of images. We can divide the gallery into $K$ groups as $\mathcal{I}=\{\mathbf{S}_k$\}, $k\in \{1,2,..., K\}$ using clustering algorithms. Here we use K-Means because of its simplicity and efficiency. For each cluster $\mathbf{S}_k$, we can calculate a centroid of the features:

\begin{equation}
	\label{equc}
	\mathbf{c}_{k}=\frac{1}{\left|\mathbf{S}_k\right|} \sum_{\mathbf{x}_{i} \in \mathbf{S}_{k}} \mathbf{x}^\mathcal{I}_{i}
\end{equation}

Given a query feature, we calculate its distances with the centroids and then rank the centroids according to this distance in ascending order. The rank of the centroids is denoted as:


\begin{equation}
	\mathcal{R}^c = (r_1, r_2, r_3, ..., r_K)
\end{equation}

Where $r_i$ stands for the index of the centroid ranked $i$. Using the rank of the centroids, we can get a ranked  image gallery $\dot{\mathcal{I}}$ grouped by clusters denoted as:
%

\begin{equation}
	\dot{\mathcal{I}}=(\mathbf{S}_{r_{1}}, \mathbf{S}_{r_{2}}, \mathbf{S}_{r_{3}}, ..., \mathbf{S}_{r_{K}})
\end{equation}

\subsubsection{Random Feature Subspace Ensembling}

Clustering in high-dimensional space has many problems. High-dimensional data often lie close to low-dimensional structures~\cite{c:102}. Many feature dimensions are irrelevant or redundant and can mask potential clusters when the data is noisy\cite{c:103}. 
The famous random decision forest algorithm~\cite{c:129} showed that combing multiple decision trees trained on randomly selected subspaces can substantially improve generalization. Many follow-up works also demonstrate its potential with other algorithms. Sampling random subspaces from the original feature space can reduce the dimension and increase the variety of clustering results.
Therefore, we introduce the technique that randomly divides features into subspaces. For example, given a feature vector $\mathbf{x} \in\mathbb{R}^D$, we can divide it into $M$ subspaces as:

\begin{equation}
	\mathbf{x}=\underbrace{\mathbf{x}^{m_1}, \ldots, \mathbf{x}^{m_{D^*}}}_{\textit{sub}_{1}(\mathbf{x})}, \ldots, \underbrace{\mathbf{x}^{m_{D-D^{*}+1}}, \ldots, \mathbf{x}^{m_D}}_{\textit{sub}_{M}(\mathbf{x})}
\end{equation}
where $\mathbf{x}^i$ denotes the $i$-th dimension of $\mathbf{x}$ , $D^*=\frac{D}{M}$ and $\{m_1, m_2, ..., m_D\}$ is a random permutation of the original channel indexes.

Split the gallery feature $\mathcal{F}^\mathcal{I}$ and perform clustering in each subspace, and we get $M\times K$ clusters with corresponding centroids, where $k\in \{1, 2, ..., K\}$ and $j\in \{1, 2, ..., M\}$. In the $j$-th subspace, we use the nearest centroid  $\mathbf{c}_i^j$ to replace $\textit{sub}_j(\mathbf{x}_i)$:
\begin{equation}
	\label{replace}
	\mathbf{\dot{x}}_i = \mathbf{c}_i^1, \mathbf{c}_i^2, ..., \mathbf{c}_i^M
\end{equation}


\subsubsection{Feature Fusion}

In the above method, dense gallery features are replaced with sparse centroids. Representing the original feature space with a few points loses fine-grained ranks within clusters, especially when $K$ is small. 
 Meanwhile, If a positive sample is classified into the wrong cluster, its retrieval result will deviate from the correct rank. In order to compensate, we use a convex combination to fuse the original features and cluster centers, producing the final feature for retrieval ranking.
\begin{equation}
	\mathbf{\ddot{x}} = (1-\lambda)\times\mathbf{\dot{x}} + \lambda\times\mathbf{x}
\end{equation}

Finally, for each query feature $\mathbf{q}$, the distance $\mathbf{d}_i$ between the query and each image is given by:
\begin{equation}
	\mathbf{d}_i=d(\mathbf{q},  \mathbf{\ddot{x}}_i)
\end{equation}
where $d(\cdot,\cdot)$ denotes euclidean distance function.

\subsection{Distribution Alignment Loss}
The Cluster-then-Retrieve method can greatly reduce the intra-class variance of gallery images, but the domain gap between sketches and images is not resolved explicitly. Inspired by~\cite{c:126},  we use Kullback--Leibler divergence to align the image and sketch features with a common Gaussian distribution. For each batch of training sample $\mathbf{X}=\{\mathbf{x}_i\in \mathbb{R}^D\}^N$, we sample a random feature batch $\mathbf{P}=\{\mathbf{p}_i\in \mathbb{R}^D\}^N$ from a given Gaussian distribution $\mathcal{N}(0,1)$. The distribution loss is formulated as:
\begin{equation}
	\mathcal{L}_\textit{dist}=KL(\mathbf{P}||E(\mathbf{X}))
\end{equation}
where $E$ is the image encoder shared between images and sketches. As this regularization is applied to both image and sketch features, the feature distribution of the two domains can be pulled closer under the guidance of Gaussian distribution, reducing the domain gap in an indirect manner.

\subsection{Network and Other Losses}
The remaining part of our model is adopted from~\cite{c:86}, where the model is initialized from a CSE-ResNet-50 network pretrained on ImageNet~\cite{c:89} and is supervised by the pretrained model through distillation during the training process. 

The main training objective is a cross-entropy loss that classifies both sketches and natural images. For each training batch $\mathbf{X}=\{\mathbf{x}_i\in \mathbb{R}^D\}^N$, the classification loss can be written as:

\begin{equation}
	\mathcal{L}_\textit{cls}=-\sum_{i=1}^{N} \log \frac{\exp \left(\alpha_{y_{i}}^{\top} \mathbf{x}_{i}+\beta_{y_{i}}\right)}{\sum_{j \in \mathcal{C}^{tr}} \exp \left(\alpha_{j}^{\top} \mathbf{x}_{i}+\beta_{j}\right)}
\end{equation}
where $\alpha$ and $\beta$ are the weight and bias of the classifier.

To preserve features learned from the ImageNet and improve the model's discriminability on unseen classes, \cite{c:86} proposes a distillation loss formulated as:

\begin{equation}
	\mathcal{L}_\textit{teacher}=-\sum_{i=1}^{N} \sum_{k \in \mathcal{C}^{T}} p_{i, k}^{\prime} \log \frac{\exp \left(\gamma_{k}^{\top} f_{i}+\delta_{k}\right)}{\sum_{j \in \mathcal{C}^{T}} \exp \left(\gamma_{j}^{\top} f_{i}+\delta_{j}\right)}
\end{equation}
where $\gamma$ and $\delta$ are the weight and bias terms in the ImageNet label classifier of the teacher model. $\mathcal{C}^{T}$ denotes the classes of the teacher model prediction. $p_{i, k}$, i.e., the soft label is given by $p_{i, k}=\textit{Softmax}(\lambda_t\cdot \mathbf{t}_i + \lambda_a \cdot \mathbf{a}_{y_i})$. Where $\mathbf{t}_i$ is the classification logit for class $i$ given by the teacher network, and $\mathbf{a}_{y_i}$ is the semantic similarity between class $i$ and label class $j$ given by WordNet~\cite{c:104}. $\lambda_t$ and $\lambda_a$ control the importance of classification logit and semantic similarity.

The full objective function of our network is:

\begin{equation}
	\mathcal{L}=\lambda_1\mathcal{L}_\textit{cls}+ \lambda_2\mathcal{L}_\textit{teacher}+\lambda_3\mathcal{L}_\textit{dist}
\end{equation}
where $\lambda_1$, $\lambda_2$ and $\lambda_3$ are hyperparameters to balance the contribution of each loss.

\section{Experiment}
\subsection{Datasets}
To verify the effectiveness of our method, we follow the common practice of previous works and experimentally evaluate our model on two popular SBIR datasets: Sketchy~\cite{c:97} and TU-Berlin~\cite{c:106}.

\subsubsection{Sketchy Dataset (Extended)}
The Sketchy dataset
contains a total of 75,471 sketches and 12,500 natural images of 125 categories. Liu et al.\cite{c:87} collect additional 60,502 natural images, which yields in total 73,002 images. We use 25 classes as the test set following the same protocol proposed by~\cite{c:81}. 

\subsubsection{TU-Berlin Dataset}
The TU-Berlin dataset~\cite{c:106} consists of 250 categories, 20,000 sketches, and 204,489 natural images provided by~\cite{c:99} . Compared to Sketchy, it suffers from a severe imbalance between sketches and images. The scarcity of sketches challenges the network's generalizability on varied hand-drawn sketches. We also follow~\cite{c:81} and randomly select 30 classes as the test set.

\subsection{Implementation Details}
We implement our model on the codebase provided by~\cite{c:86}. The SE-ResNet-50 network pre-trained on ImageNet is chosen to initialize the teacher and student networks. We train our model with the Adam optimizer with $\beta_1=0.9$, $\beta_2=0.999$. The learning rate is initialized as $1e-4$ and exponentially decayed to $1e-7$. We set $\lambda_1$ to $1$, $\lambda_2$ to $1$ and $\lambda_3$ to $0.1$ unless stated. $\lambda_t$ and $\lambda_a$ are set to $1$ and $0.3$ for all experiments following the setting of~\cite{c:86}. The semantic similarity is given by the WordNet python interface from nltk corpus reader. The random feature batch in our distribution alignment loss follows the distribution of $\mathcal{N}(0,1)$. We set the batch size to 80 and train the model for 20 epochs.

For the hyperparameters of clustering, we set $M$ to $2$ and $K$ to $32$ for all experiments unless otherwise stated. Fuse coefficient $\lambda$ is set to 0.2. The effects of different hyperparameters will be discussed in the appendix.

\subsection{Comparison with Existing Methods}

\begin{table*} \centering
    \caption{Retrieval performance comparison. $\dagger$ denotes results obtained using binary codes. SAKE (reimp.) is the reimplemented version of SAKE trained with batch size of 80. SAKE w/ $\mathcal{L}_{dist}$ is SAKE with distribution alignment loss. The best results are in bold.}
    \label{results}
    \begin{tabular}{ccccccc}
    \hline
    & \multirow{2}{*}{Methods} & \multirow{2}{*}{Dim} & \multicolumn{2}{c}{Sketchy} & \multicolumn{2}{c}{TU-Berlin} \\ \cline{4-7} 
                             &                                    &              & mAP@all & Prec@100      & mAP@all & Prec@100\\ \hline
    &SEM-PCYC (CVPR’2019)           & $64\dagger$  & 0.344    & 0.399         & 0.293      & 0.392   \\
    & SAKE (ICCV’2019)        & $64\dagger$           & {0.364}    & 0.487         & {0.359}      & 0.481   \\
    &SEM-PCYC (CVPR’2019)          & 64           & 0.349    & 0.463         & 0.297      & 0.426   \\
    & CSDB (BMVC'2019)                              & 64           & {0.376}    & {0.484}         & 0.254      & 0.355   \\
    & TVT (AAAI'2022)                                      & $64\dagger$          & 0.553      & \textbf{0.727} & 0.396   & \textbf{0.606} \\
    & ClusterRetri (Ours)                 & $64\dagger$  & 0.531    & {0.465}         & \textbf{0.439}      & 0.401  \\
    & ClusterRetri (Ours)                 & 64  & \textbf{0.585}    & \textbf{0.618}         & \textbf{0.498}      & \textbf{0.529}  \\ \cline{2-7}
    & SAKE (ICCV'2019)        & 512          & 0.547    & 0.692         & 0.475      & 0.599   \\
    & SAKE (reimp.)                              & 512          & 0.582    & 0.706         & 0.428      & 0.346   \\
    & DSN (IJCAI'2021)                               & 512          & 0.583    & 0.704         & 0.481      & 0.586   \\
    & RPKD (ACMMM'2021)                              & 512          & 0.613    & 0.723         & 0.486      & \textbf{0.612}   \\
    & NAVE (IJCAI'2021)                              & 512          & 0.613    & 0.725         & 0.493      & 0.607   \\
    & TVT (AAAI'2022)                                      & 384                  & 0.648       & \textbf{0.796}         & 0.484     & \textbf{0.662}\\
    & SAKE w/ $\mathcal{L}_{dist}$ (Ours)                       & 512          & 0.613    & 0.730         & 0.498      & 0.500   \\
    & ClusterRetri (Ours)                 & $512\dagger$ & \textbf{0.781}    & \textbf{0.764}         & \textbf{0.672}      & \textbf{0.637}  \\
    & ClusterRetri (Ours)                 & 512          & \textbf{0.762}    & 0.760         & \textbf{0.597}      & 0.584 \\ \hline
    \end{tabular}
    \end{table*}

We compare our ClusterRetri model with various methods of ZS-SBIR, including SEM-PCYC~\cite{c:101}, SAKE~\cite{c:86}, CSDB~\cite{c:113}, DSN~\cite{c:84} and TVT ~\cite{c:130}. The results are shown in Table \ref{results}. We first retrain the baseline model, i.e., the SAKE~\cite{c:86}. It should be noted that with batch size of 80, SAKE can achieve 0.582 mAP@all on Sketchy dataset, which surpasses the result in the original paper, so we train our models with batch size of 80. 

With the proposed feature distribution loss, our baseline model can achieve 0.613 mAP@all on Sketchy dataset and 0.498 mAP@all on TU-Berlin dataset, which is comparable with state-of-the-art methods. 

Apply our Cluster-then-Retrieve method to the feature generated by the distribution regulated network, and we can have a huge boost on the retrieval performance. Specifically, in 512-d real value retrieval, our method achieves 0.762 and 0.597 mAP@all on Sketchy and TU-Berlin, respectively. 

Following previous works~\cite{c:101,c:86}, we use iterative quantization~\cite{c:88} (ITQ) to generate binary features. Here the input of the ITQ algorithm is the reconstructed feature using clustered centroids. Specifically, the features' subvectors are replaced with the closest centroids in corresponding subspaces. Then we calculate the hamming distance to conduct retrieval. Surprisingly, the performance is further improved, achieving 0.781 mAP@all on Sketchy dataset and 0.672 mAP@all on TU-Berlin dataset. We assume this is probably because the iterative quantization can reduce redundant features that may affect retrieval. 

Using 64-d features, our method's performance is comparable with the previous SOTA method with 512-d features, demonstrating our ClusterRetri's effectiveness. Under the 64-d binary setting, the real value features are compact enough, so our method experiences a drop in performance, but it also outperforms other methods by a large margin.


\section{Discussion}

\subsection{Qualitative Analysis}

\subsubsection{Example of Retrievals}

\begin{figure}[h]
    \centering
    \includegraphics[width=0.45\textwidth]{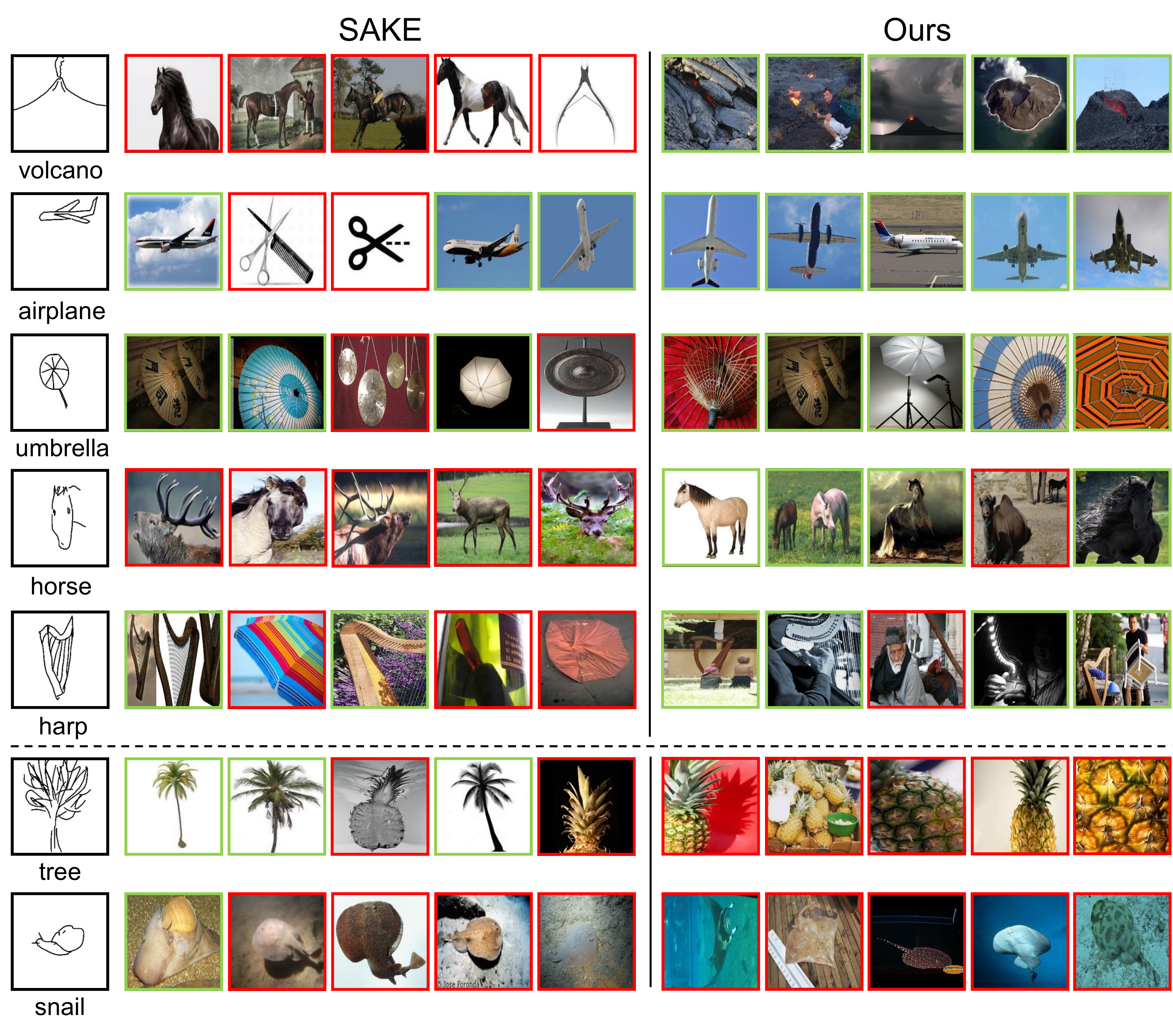}
    \caption{Comparison of top-5 images obtained by SAKE and our ClusterRetri on Sketchy dataset. The experiment is conducted using 512-d feature. Mis-retrieved images are marked with red borders. The last two rows are representative failure cases. It can be seen that through clustering, our method can retrieve using not only outline similarities but also color and texture features.}
    \label{fig:examples}
\end{figure}

We visualize several typical results on Sketchy by our ClusterRetri and our baseline SAKE for qualitative analysis in Fig. \ref{fig:examples}. The first column of sketches in Fig. \ref{fig:examples} are queries and their class labels. To their right are the top 5 images retrieved by baseline and our method. In both experiments, the models are trained with 512-d real value features. The images with green borders are positive samples, while those with red are negative ones. 

Through comparison, we can find the superiority of our method in that we can retrieve images that are not similar in shapes. For example, in the first row of Fig. \ref{fig:examples}, our method can find volcano images that are very dissimilar to the query sketch because clustering can group the images with the color of flame and the texture of rocks together. Another example is in the $2$-nd row, where the baseline mis-retrieved the scissors as planes because of the visual similarity of edges. Our method, however, can prevent such outliers once the clustering is successful. This observation confirms the effectiveness of leveraging color and texture information through clustering. However, our method also has drawbacks. In the last two rows, due to the poor quality of the query sketches, they are classified into the wrong clusters, negatively impacting the clustering performance.

\subsubsection{Effectiveness of Domain Adaptation}
\begin{figure}[h]
    \centering
    \subfigure[All Pairs KL Loss]{\includegraphics[width=0.233\textwidth]{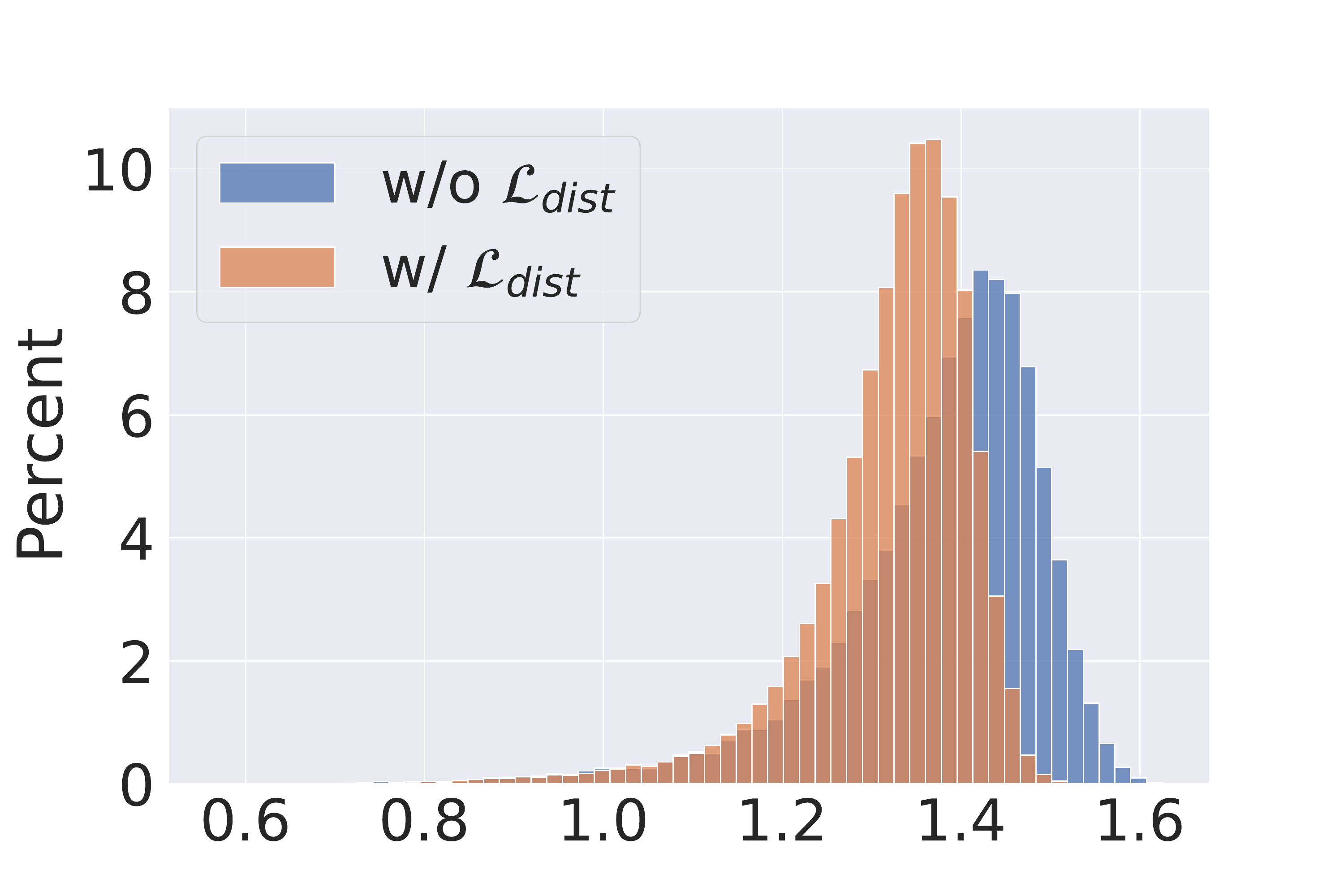}}
    \subfigure[Positive Pairs KL Loss]{\includegraphics[width=0.233\textwidth]{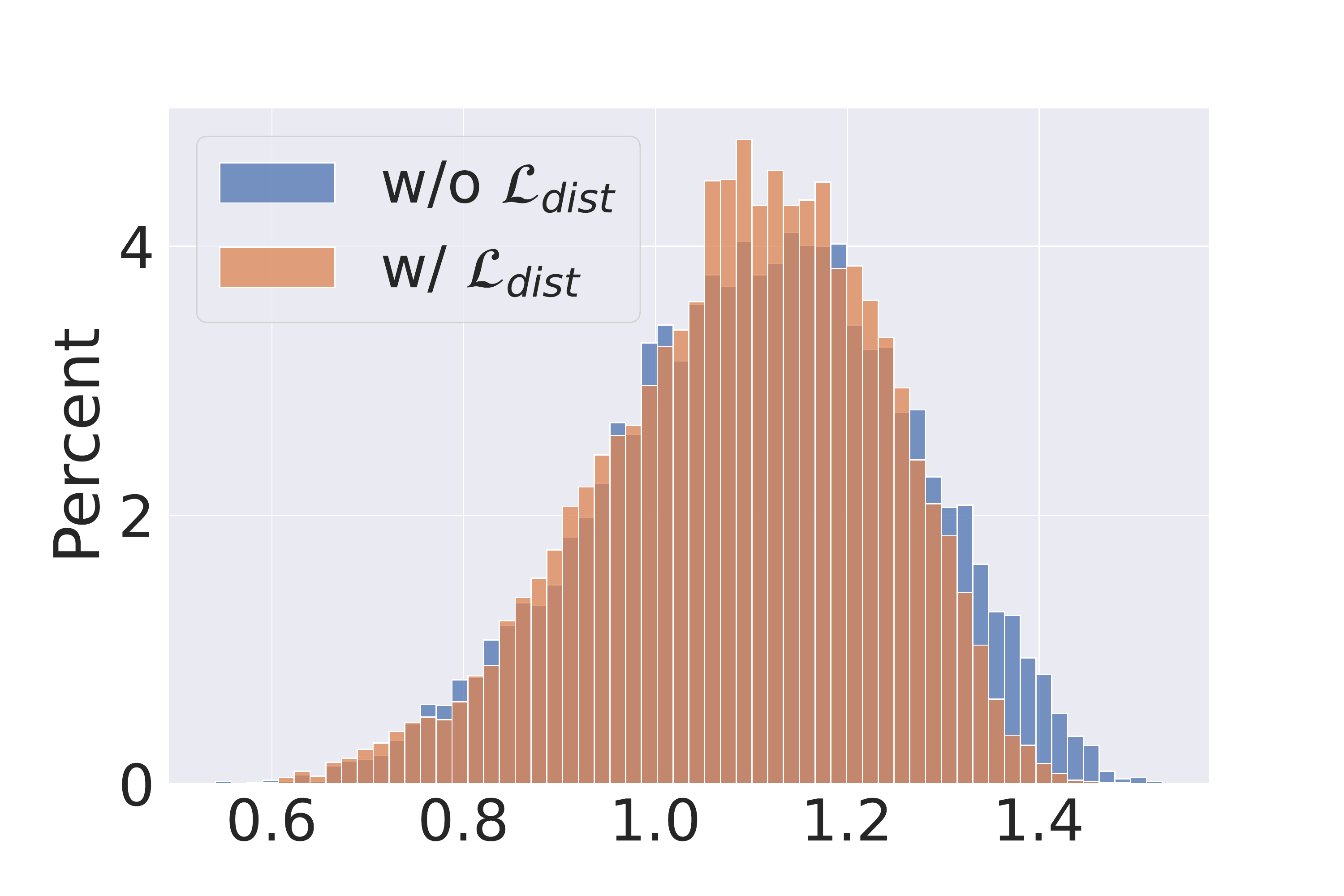}}
    \subfigure[All Pairs Clustering]{\includegraphics[width=0.233\textwidth]{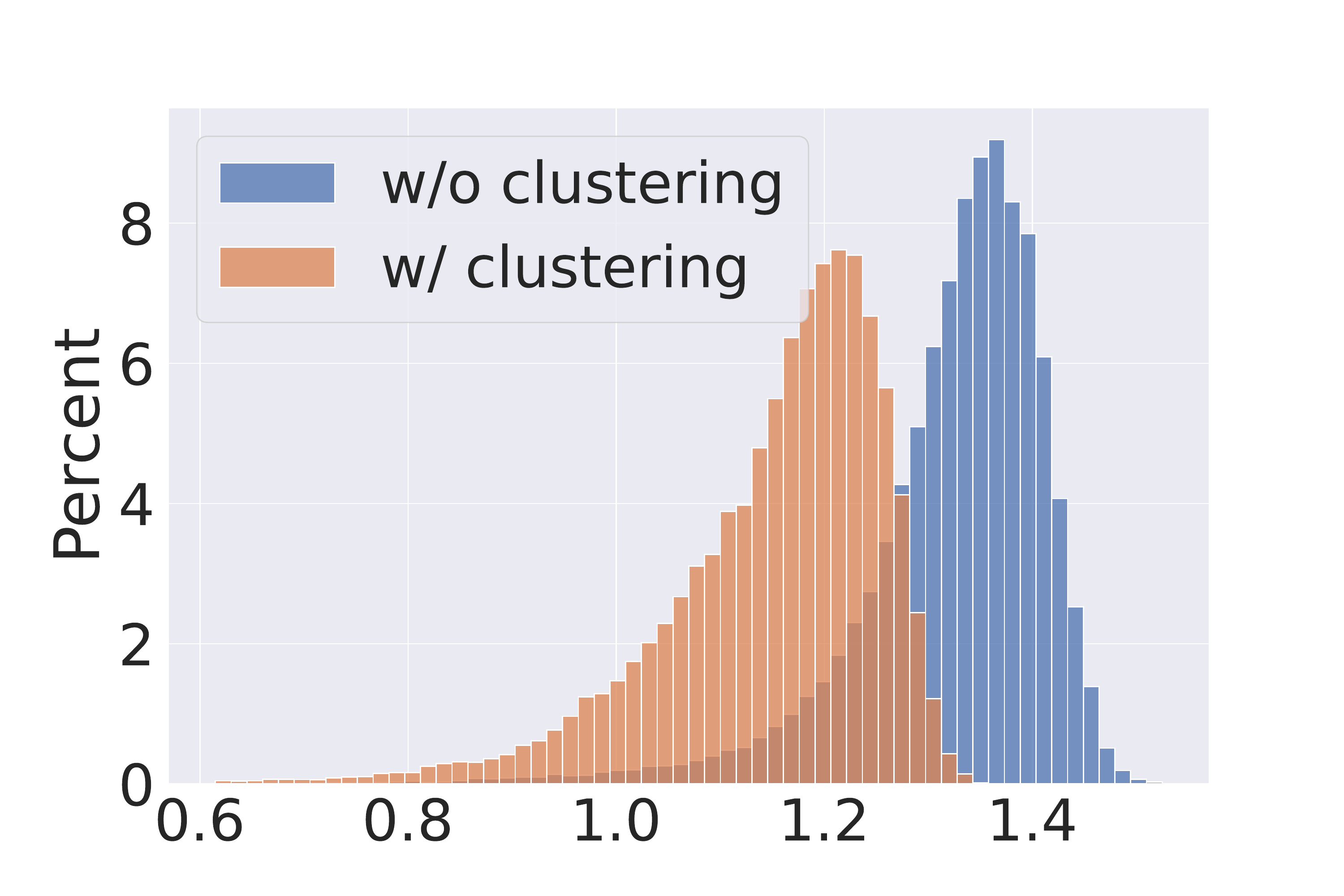}}
    \subfigure[Positive Pairs Clustering]{\includegraphics[width=0.233\textwidth]{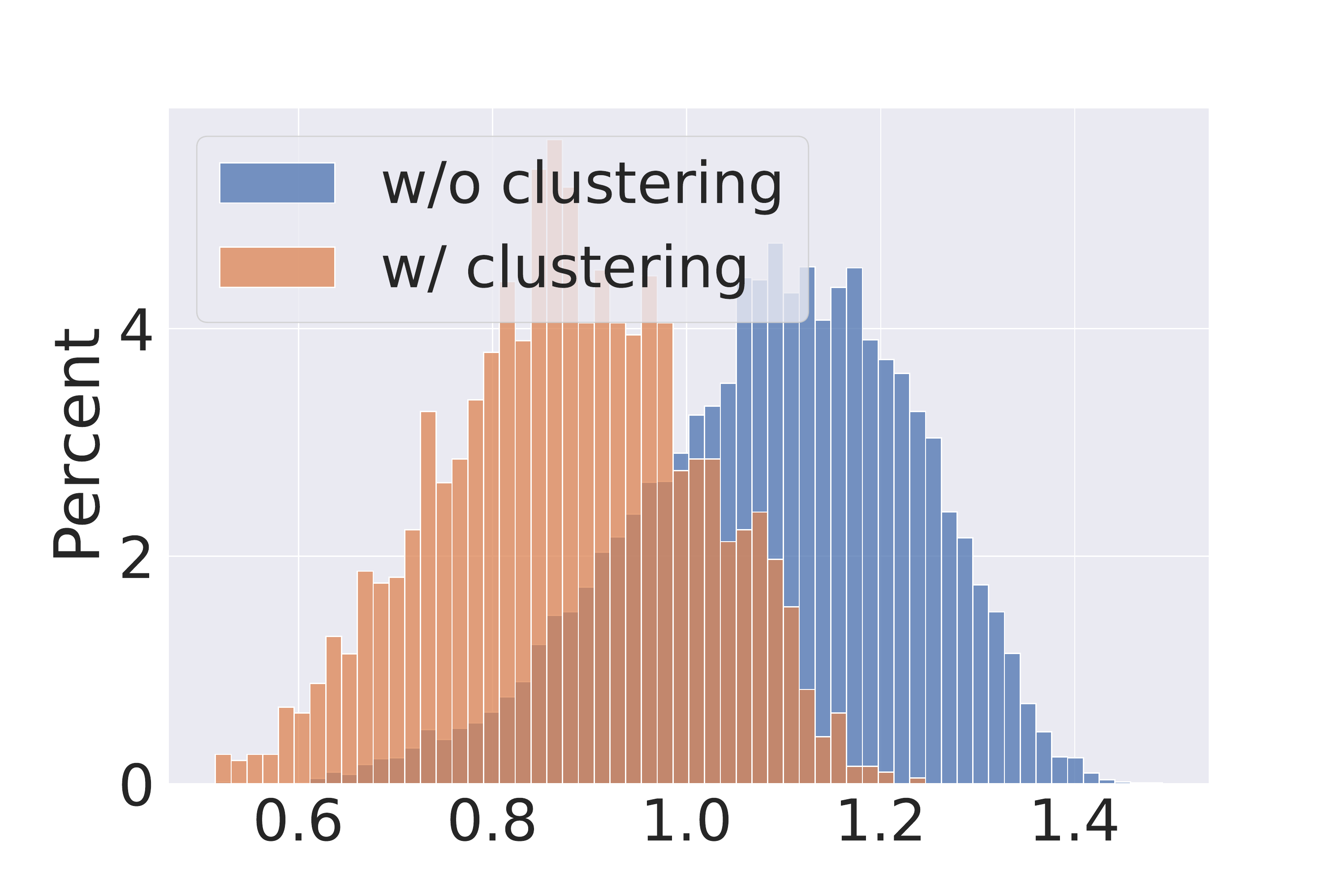}}
    \caption{Histogram of the distances between sketch and image domains. (a) and (c) illustrates distances between all sketch-image pairs, (b) and (d) illustrates distances between sketches and images from the same category.}
    \label{fig:dist}
\end{figure}

We use distance histograms to visualize the distance of the two domains. We calculate the distances between all possible sketch-image pairs in the test set of Sketchy dataset, then sample 9000 negative pairs and 1000 positive pairs to plot Fig. \ref{fig:dist}. It can be seen from Fig. \ref{fig:dist} (a) and (b) that with the distribution alignment loss, the overall distances between sketches and images are reduced, indicating the effectiveness of domain adaptation. Moreover, as Fig.\ref{fig:dist} (c) and (d) show, with the cluster-rectified feature, the distances between sketch queries and images are also greatly reduced. Using the cluster centers as prototypes to replace the original features can filter out noises such as irrelevant background or multiple objects, facilitating the distance comparison between images and sparse sketches.

\subsection{Quantitative Analysis}
\subsubsection{Clustering}
\label{section:clustering}
\begin{figure}
    \centering
    \subfigure[Influence of K]{\includegraphics[width=0.225\textwidth]{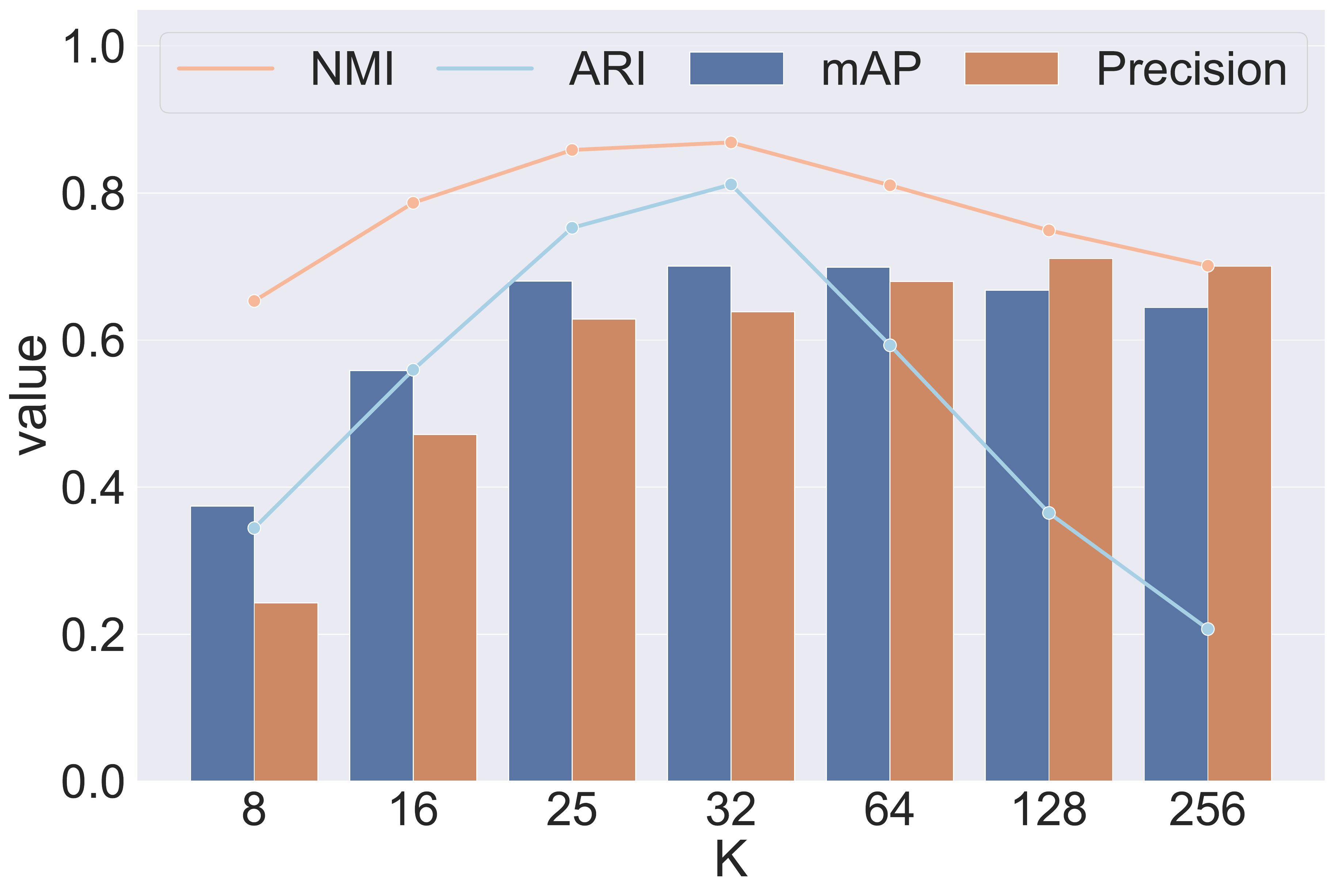}}
    \subfigure[Influence of M]{\includegraphics[width=0.225\textwidth]{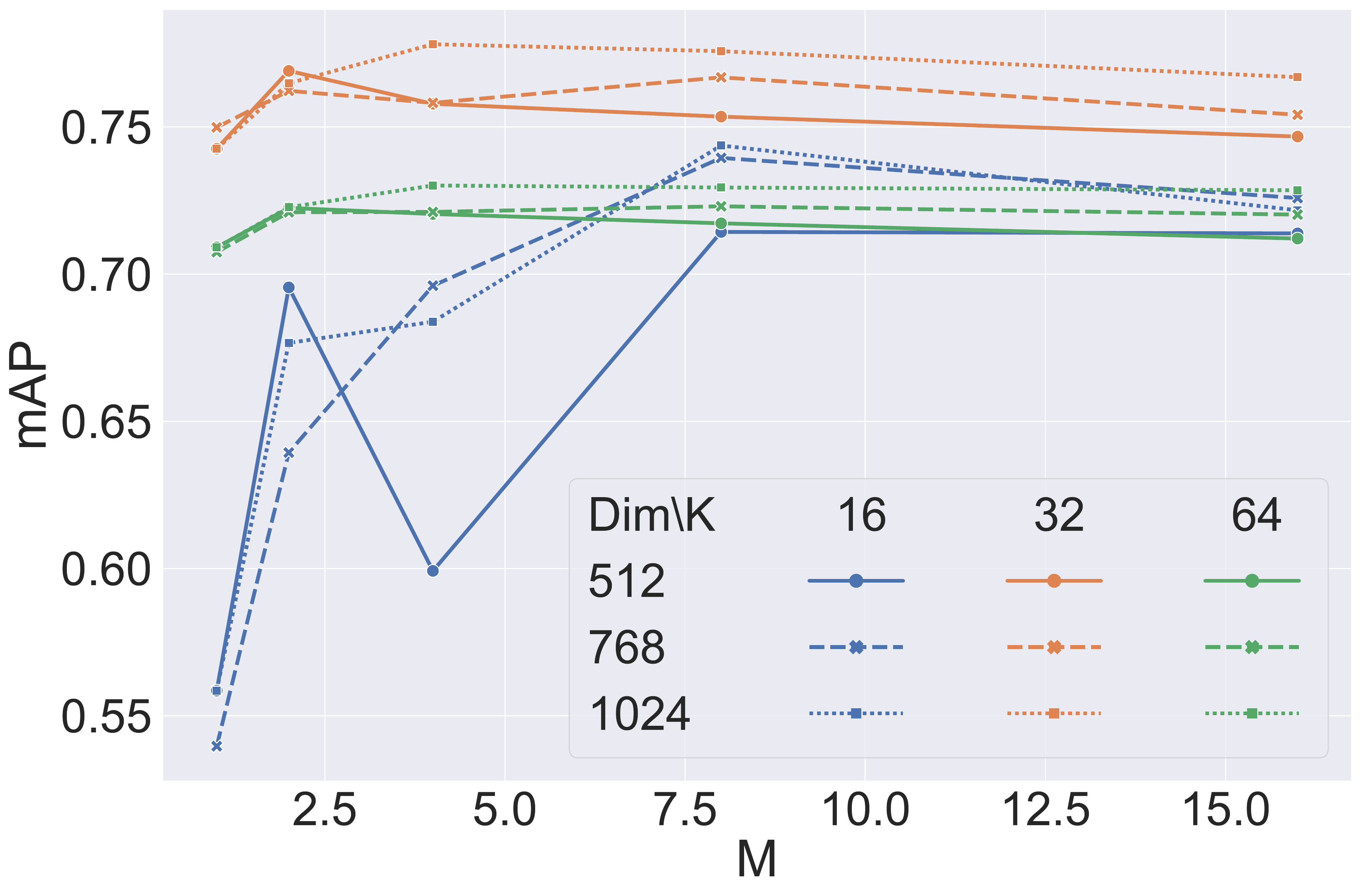}}
    \caption{The influence of parameters on Sketchy dataset using 512-d real value features. The settings of $Dim > 512$ in (b) are obtained by sampling extra random subspaces from the original 512-d features.}
    \label{fig:mk}
\end{figure}

Here we experimentally validate the effectiveness of our Cluster-then-Retrieve method. We first perform K-Means clustering on the 512-d gallery features generated by our baseline model and use the clusters to perform retrieval. The results are illustrated in Fig. \ref{fig:mk}. Here we use normalized mutual information (NMI)~\cite{c:110} and adjusted rand score (ARI)~\cite{c:111} to measure the performance of clustering. It can be seen that there is a strong correlation between clustering performance and retrieval mAP. 
This suggests that adopting better clustering algorithms in our pipeline can further improve retrieval performance. It should be noted that ideally, the best choice of K should be equal to the class number of gallery images. However, the method performs best when $K$ is around $30$, which is slightly higher than $25$ classes in the test set of Sketchy. This can be explained by intra-class variance. Multiple small clusters may exist in a class of feature points, so grouping all the images into one cluster may affect the estimation of the feature center.


\subsubsection{Random Subspace Ensemble}
We then visualize the effect of the number of subspaces on retrieval performance in Fig. \ref{fig:mk}. We can see a large improvement when $M$ is set from $1$ to $2$. When we directly use the K-Means centroids as proxies for retrieval, i.e., $M=1$, retrieval performance heavily depends on the clustering quality of one shot. If a sample is misclassified to a cluster it does not belong to, it can hardly be placed in the proper rank. Dividing subspace softens this risk, as the distance is calculated in different subspaces, and its rank is decided through ensembled results of all the subspaces. Noted that in Fig.\ref{fig:mk}, the settings $Dim = 768$ and $Dim = 1024$ show that the ensemble results can be further improved by sampling extra random subspaces from the 512-d features.

	
\subsection{Analysis of Efficiency Improvement}

\begin{table}[h]
\centering
 \caption{Efficiency Comparison on Sketchy dataset}
 \label{table:eff}
\setlength\tabcolsep{2pt}
\begin{tabular}{ccccccc}
\hline
Method                    & Dim & M & K  & Time(s) & mAP@all & Prec@100 \\ \hline
\multirow{2}{*}{Baseline} & 512 & - & -  & 82.7  & 0.582   & 0.706    \\
                          & 64  & - & -  & 9.1   & 0.364   & 0.487    \\
\multirow{2}{*}{Ours}     & 512 & 1 & 32 & \textbf{2.6}   & 0.743   & 0.711    \\
                          & 512 & 2 & 32 & 10.8  & \textbf{0.769}   & \textbf{0.742}    \\ \hline
\end{tabular}
\end{table}

Our gallery feature clustering pipeline can achieve impressive results. It should be noted that the K-Means clustering and subspace division is very similar to the classical vector quantization (VQ)~\cite{c:127} and product quantization (PQ)~\cite{c:123} algorithms in information retrieval, which focus on improving the retrieval efficiency by lookup tables.

While driven by different motivations, our method can also enjoy the speed improvement brought by PQ.
We conduct experiments on a machine using Intel Xeon Silver 4215R. Table.\ref{table:eff} shows the time usage of distance calculation and corresponding retrieval performance. The distance size is $15229 \times 17101$. It can be seen that our method can achieve superior performance and efficiency simultaneously. Using 512-d features and ClusterRetri, the time of distance calculation can be much faster than that of 64-d real number features.


\section{Ablation Study}
\begin{table}[h]
	\centering
    \caption{Ablation of different modules. Noted that mAP stands for mAP@all and Prec stands for Prec@100.}
    \label{ablation}
    \setlength\tabcolsep{2pt}
    \begin{tabular}{cccccccc}
    \hline
    \multirow{2}{*}{KMeans} & \multirow{2}{*}{Fuse} & \multirow{2}{*}{Subspace} & \multirow{2}{*}{$\mathcal{L}_{dist}$} & \multicolumn{2}{c}{Sketchy} & \multicolumn{2}{c}{TU-berlin} \\ \cline{5-8} 
                            &                            &                           &                          & mAP& Prec           & mAP & Prec            \\ \hline
                            &                            &                           &                          & 0.582  & 0.706              & 0.428   & 0.346               \\
    \checkmark              &                            &                           &                          & 0.701  & 0.639              & 0.520   & 0.471               \\
    \checkmark              & \checkmark                 &                           &                          & 0.706  & 0.700              & 0.528   & 0.524               \\
    \checkmark              &                            & \checkmark                &                          & 0.727  & 0.683              & 0.511   & 0.473               \\
    \checkmark              & \checkmark                 & \checkmark                &                          & 0.732  & 0.697              & 0.517   & 0.490               \\
                            &                            &                           & \checkmark               & 0.613  & 0.730              & 0.498   & 0.500               \\
    \checkmark              &                            &                           & \checkmark               & 0.743  & 0.711              & 0.576   & 0.530               \\
    \checkmark              & \checkmark                 &                           & \checkmark               & 0.746  & 0.742              & \textbf{0.594}   & \textbf{0.600}               \\
    \checkmark              &                            & \checkmark                & \checkmark               & \textbf{0.769}  & 0.742              & 0.583   & 0.532               \\
    \checkmark              & \checkmark                 & \checkmark                & \checkmark               & 0.761  & \textbf{0.760}   & 0.587   & 0.574               \\ \hline
    \end{tabular}
    \end{table}

Here we evaluate the effectiveness of each component of our method. The baseline model is adopted from~\cite{c:86} and trained using 512 dimension features with batch size of 80. We gradually add clustering, reranking, and subspace division to the pipeline. Then we use the model trained with distribution alignment loss and repeat the above experiments to demonstrate its effectiveness. Note that the experiments are performed using $K=32$ and $M=2$.

It can be seen from Table \ref{ablation} that Simply choosing an appropriate $K$ for K-Means clustering is enough to get a huge boost on the performance. Specifically, we can achieve an mAP of $0.701$ when $K=32$, outperforming any SOTA algorithms by a large margin. Clustering combined with feature fusion or subspace division further improves the result.  

Distribution alignment loss alone can improve the baseline by $5\%$ and $16\%$ on Sketchy and TU-berlin. It can also provide 0.042 and 0.059 extra improvements when combined with K-Means clustering. This shows the potential of reducing intra-class variance and domain gap simultaneously.

Finally, our full pipeline can improve the mAP@all by $31\%$ and $39\%$ on Sketchy and TU-Berlin datasets.

\section{Conclusion}
This paper tackles the problem of zero-shot sketch-based image retrieval from a new perspective, i.e., by leveraging the gallery information. Domain gap and intra-class variance are inherent problems of ZS-SBIR. Inferring from sparse geometry information alone is inadequate for the varied natural images. Therefore, we adopt feature clustering to utilize the relationship between gallery images. The pre-processing technique can effectively use color and texture features that sketches lack, giving a huge boost to existing ZS-SBIR methods. 

The proposed pipeline is flexible, as clustering can be conducted on features generated by any algorithm. State-of-the-art clustering algorithms can also be adopted to improve the overall performance further. Moreover, the idea of Cluster-then-Retrieve has great potential on other cross-modal category-level retrieval tasks. 

\bibliography{aaai23.bib}

\begin{thebibliography}{42}
\providecommand{\natexlab}[1]{#1}

\bibitem[{Deng et~al.(2009)Deng, Dong, Socher, Li, Li, and Fei-Fei}]{c:89}
Deng, J.; Dong, W.; Socher, R.; Li, L.-J.; Li, K.; and Fei-Fei, L. 2009.
\newblock Imagenet: A large-scale hierarchical image database.
\newblock In \emph{2009 IEEE Conference on Computer Vision and Pattern
  Recognition (CVPR)}, 248--255. IEEE.

\bibitem[{Dutta and Akata(2019)}]{c:101}
Dutta, A.; and Akata, Z. 2019.
\newblock Semantically Tied Paired Cycle Consistency for Zero-Shot Sketch-Based
  Image Retrieval.
\newblock In \emph{Computer Vision and Pattern Recognition}.

\bibitem[{Dutta and Biswas(2019)}]{c:113}
Dutta, T.; and Biswas, S. 2019.
\newblock Style-Guided Zero-Shot Sketch-based Image Retrieval.
\newblock In \emph{BMVC}.

\bibitem[{Eitz, Hays, and Alexa(2012)}]{c:106}
Eitz, M.; Hays, J.; and Alexa, M. 2012.
\newblock How do humans sketch objects.
\newblock In \emph{International Conference on Computer Graphics and
  Interactive Techniques}.

\bibitem[{Eitz et~al.(2010)Eitz, Hildebrand, Boubekeur, and Alexa}]{c:94}
Eitz, M.; Hildebrand, K.; Boubekeur, T.; and Alexa, M. 2010.
\newblock An evaluation of descriptors for large-scale image retrieval from
  sketched feature lines.
\newblock \emph{Comput. Graph.}, 34: 482--498.

\bibitem[{Eitz et~al.(2011)Eitz, Hildebrand, Boubekeur, and Alexa}]{c:92}
Eitz, M.; Hildebrand, K.; Boubekeur, T.; and Alexa, M. 2011.
\newblock Sketch-Based Image Retrieval: Benchmark and Bag-of-Features
  Descriptors.
\newblock \emph{IEEE Transactions on Visualization and Computer Graphics}, 17:
  1624--1636.

\bibitem[{Elhamifar and Vidal(2013)}]{c:102}
Elhamifar, E.; and Vidal, R. 2013.
\newblock Sparse Subspace Clustering: Algorithm, Theory, and Applications.
\newblock \emph{IEEE Transactions on Pattern Analysis and Machine
  Intelligence}, 35: 2765--2781.

\bibitem[{Ge et~al.(2013)Ge, He, Ke, and Sun}]{c:124}
Ge, T.; He, K.; Ke, Q.; and Sun, J. 2013.
\newblock Optimized Product Quantization for Approximate Nearest Neighbor
  Search.
\newblock \emph{2013 IEEE Conference on Computer Vision and Pattern
  Recognition}, 2946--2953.

\bibitem[{Gong et~al.(2012)Gong, Lazebnik, Gordo, and Perronnin}]{c:88}
Gong, Y.; Lazebnik, S.; Gordo, A.; and Perronnin, F. 2012.
\newblock Iterative quantization: A procrustean approach to learning binary
  codes for large-scale image retrieval.
\newblock \emph{IEEE transactions on pattern analysis and machine
  intelligence}, 35(12): 2916--2929.

\bibitem[{Gray(1984)}]{c:127}
Gray, R. 1984.
\newblock Vector quantization.
\newblock \emph{IEEE ASSP Magazine}, 1(2): 4--29.

\bibitem[{Guo et~al.(2017)Guo, Liu, Wang, Luo, Wen, and Lu}]{c:82}
Guo, L.; Liu, J.; Wang, Y.; Luo, Z.; Wen, W.; and Lu, H. 2017.
\newblock Sketch-based Image Retrieval using Generative Adversarial Networks.
\newblock \emph{Proceedings of the 25th ACM international conference on
  Multimedia}.

\bibitem[{Hearst and Pedersen(1996)}]{c:119}
Hearst, M.~A.; and Pedersen, J.~O. 1996.
\newblock Reexamining the cluster hypothesis: scatter/gather on retrieval
  results.
\newblock In \emph{SIGIR '96}.

\bibitem[{Ho(1998)}]{c:129}
Ho, T.~K. 1998.
\newblock The Random Subspace Method for Constructing Decision Forests.
\newblock \emph{{IEEE} Trans. Pattern Anal. Mach. Intell.}, 20(8): 832--844.

\bibitem[{Hu and Collomosse(2013)}]{c:91}
Hu, R.; and Collomosse, J.~P. 2013.
\newblock A performance evaluation of gradient field HOG descriptor for sketch
  based image retrieval.
\newblock \emph{Comput. Vis. Image Underst.}, 117: 790--806.

\bibitem[{Hu, Wang, and Collomosse(2011)}]{c:93}
Hu, R.; Wang, T.; and Collomosse, J.~P. 2011.
\newblock A bag-of-regions approach to sketch-based image retrieval.
\newblock \emph{2011 18th IEEE International Conference on Image Processing},
  3661--3664.

\bibitem[{Huang et~al.(2021)Huang, Sun, Han, Gao, and Sang}]{c:85}
Huang, Z.; Sun, Y.; Han, C.; Gao, C.; and Sang, N. 2021.
\newblock Modality-Aware Triplet Hard Mining for Zero-shot Sketch-Based Image
  Retrieval.
\newblock \emph{arXiv preprint arXiv:2112.07966}.

\bibitem[{Hubert and Arabie(1985)}]{c:111}
Hubert, L.~J.; and Arabie, P. 1985.
\newblock Comparing partitions.
\newblock \emph{Journal of Classification}, 2: 193--218.

\bibitem[{Jardine and van Rijsbergen(1971)}]{c:117}
Jardine, N.; and van Rijsbergen, C.~J. 1971.
\newblock The use of hierarchic clustering in information retrieval.
\newblock \emph{Inf. Storage Retr.}, 7: 217--240.

\bibitem[{J{\'e}gou, Douze, and Schmid(2011)}]{c:123}
J{\'e}gou, H.; Douze, M.; and Schmid, C. 2011.
\newblock Product Quantization for Nearest Neighbor Search.
\newblock \emph{IEEE Transactions on Pattern Analysis and Machine
  Intelligence}, 33: 117--128.

\bibitem[{Kalantidis and Avrithis(2014)}]{c:125}
Kalantidis, Y.; and Avrithis, Y. 2014.
\newblock Locally Optimized Product Quantization for Approximate Nearest
  Neighbor Search.
\newblock \emph{2014 IEEE Conference on Computer Vision and Pattern
  Recognition}, 2329--2336.

\bibitem[{Liu et~al.(2017)Liu, Shen, Shen, Liu, and Shao}]{c:87}
Liu, L.; Shen, F.; Shen, Y.; Liu, X.; and Shao, L. 2017.
\newblock Deep Sketch Hashing: Fast Free-Hand Sketch-Based Image Retrieval.
\newblock \emph{2017 IEEE Conference on Computer Vision and Pattern Recognition
  (CVPR)}, 2298--2307.

\bibitem[{Liu et~al.(2019)Liu, Xie, Wang, and Yuille}]{c:86}
Liu, Q.; Xie, L.; Wang, H.; and Yuille, A.~L. 2019.
\newblock Semantic-aware knowledge preservation for zero-shot sketch-based
  image retrieval.
\newblock In \emph{Proceedings of the IEEE/CVF International Conference on
  Computer Vision}, 3662--3671.

\bibitem[{Miller(1992)}]{c:104}
Miller, G.~A. 1992.
\newblock WordNet: a lexical database for English.
\newblock In \emph{Human Language Technology}.

\bibitem[{Parsons, Haque, and Liu(2004)}]{c:103}
Parsons, L.; Haque, E.; and Liu, H. 2004.
\newblock Subspace clustering for high dimensional data: a review.
\newblock \emph{SIGKDD Explor.}, 6: 90--105.

\bibitem[{Qi et~al.(2016)Qi, Song, Zhang, and Liu}]{c:98}
Qi, Y.; Song, Y.-Z.; Zhang, H.; and Liu, J. 2016.
\newblock Sketch-based image retrieval via Siamese convolutional neural
  network.
\newblock In \emph{International Conference on Image Processing}.

\bibitem[{Saavedra, Barrios, and Orand(2015)}]{c:90}
Saavedra, J.~M.; Barrios, J.~M.; and Orand, S. 2015.
\newblock Sketch based Image Retrieval using Learned KeyShapes (LKS).
\newblock In \emph{BMVC}, volume~1, 7.

\bibitem[{Sangkloy et~al.(2016)Sangkloy, Burnell, Ham, and Hays}]{c:97}
Sangkloy, P.; Burnell, N.; Ham, C.; and Hays, J. 2016.
\newblock The sketchy database: learning to retrieve badly drawn bunnies.
\newblock In \emph{International Conference on Computer Graphics and
  Interactive Techniques}.

\bibitem[{Shaw, Burgin, and Howell(1997)}]{c:121}
Shaw, W.~M.; Burgin, R.; and Howell, P. 1997.
\newblock Performance Standards and Evaluations in IR Test Collections:
  Cluster-Based Retrieval Models.
\newblock \emph{Inf. Process. Manag.}, 33: 1--14.

\bibitem[{Shen et~al.(2018)Shen, Liu, Shen, and Shao}]{c:81}
Shen, Y.; Liu, L.; Shen, F.; and Shao, L. 2018.
\newblock Zero-Shot Sketch-Image Hashing.
\newblock \emph{2018 IEEE/CVF Conference on Computer Vision and Pattern
  Recognition}, 3598--3607.

\bibitem[{Song et~al.(2017)Song, Yu, Song, Xiang, and Hospedales}]{c:96}
Song, J.; Yu, Q.; Song, Y.-Z.; Xiang, T.; and Hospedales, T.~M. 2017.
\newblock Deep Spatial-Semantic Attention for Fine-Grained Sketch-Based Image
  Retrieval.
\newblock In \emph{International Conference on Computer Vision}.

\bibitem[{Strehl and Ghosh(2002)}]{c:110}
Strehl, A.; and Ghosh, J. 2002.
\newblock Cluster Ensembles --- A Knowledge Reuse Framework for Combining
  Multiple Partitions.
\newblock \emph{J. Mach. Learn. Res.}, 3: 583--617.

\bibitem[{Sze(2004)}]{c:122}
Sze, R. 2004.
\newblock \emph{Cluster-based information retrieval modeling}.
\newblock Ph.D. thesis, University of British Columbia.

\bibitem[{Tian et~al.(2022)Tian, Xu, Shen, Yang, and Shen}]{c:130}
Tian, J.; Xu, X.; Shen, F.; Yang, Y.; and Shen, H.~T. 2022.
\newblock {TVT:} Three-Way Vision Transformer through Multi-Modal Hypersphere
  Learning for Zero-Shot Sketch-Based Image Retrieval.
\newblock In \emph{Thirty-Sixth {AAAI} Conference on Artificial Intelligence,
  {AAAI} 2022, Thirty-Fourth Conference on Innovative Applications of
  Artificial Intelligence, {IAAI} 2022, The Twelveth Symposium on Educational
  Advances in Artificial Intelligence, {EAAI} 2022 Virtual Event, February 22 -
  March 1, 2022}, 2370--2378. {AAAI} Press.

\bibitem[{Tombros, Villa, and van Rijsbergen(2002)}]{c:120}
Tombros, A.; Villa, R.; and van Rijsbergen, C.~J. 2002.
\newblock The effectiveness of query-specific hierarchic clustering in
  information retrieval.
\newblock \emph{Inf. Process. Manag.}, 38: 559--582.

\bibitem[{Vdorhees(2017)}]{c:118}
Vdorhees, E.~M. 2017.
\newblock The Cluster Hypothesis Revisited.
\newblock \emph{SIGIR Forum}, 51: 35--43.

\bibitem[{Wang et~al.(2021{\natexlab{a}})Wang, Chen, Lin, Sigal, and
  de~Silva}]{c:126}
Wang, J.; Chen, J.; Lin, J.; Sigal, L.; and de~Silva, C.~W. 2021{\natexlab{a}}.
\newblock Discriminative feature alignment: Improving transferability of
  unsupervised domain adaptation by Gaussian-guided latent alignment.
\newblock \emph{Pattern Recognition}, 116: 107943.

\bibitem[{Wang et~al.(2021{\natexlab{b}})Wang, Wang, Yan, Wu, and Deng}]{c:84}
Wang, Z.; Wang, H.; Yan, J.; Wu, A.; and Deng, C. 2021{\natexlab{b}}.
\newblock Domain-Smoothing Network for Zero-Shot Sketch-Based Image Retrieval.
\newblock \emph{IJCAI}, 1143--1149.

\bibitem[{Yelamarthi et~al.(2018)Yelamarthi, Reddy, Mishra, and Mittal}]{c:80}
Yelamarthi, S.~K.; Reddy, S.~K.; Mishra, A.; and Mittal, A. 2018.
\newblock A zero-shot framework for sketch-based image retrieval.
\newblock In \emph{Proceedings of the European Conference on Computer Vision
  (ECCV)}, 300--317.

\bibitem[{Yu et~al.(2016{\natexlab{a}})Yu, Liu, Song, Xiang, Hospedales, and
  Loy}]{c:100}
Yu, Q.; Liu, F.; Song, Y.-Z.; Xiang, T.; Hospedales, T.~M.; and Loy, C.~C.
  2016{\natexlab{a}}.
\newblock Sketch Me That Shoe.
\newblock In \emph{Computer Vision and Pattern Recognition}.

\bibitem[{Yu et~al.(2016{\natexlab{b}})Yu, Yang, Liu, Song, Xiang, and
  Hospedales}]{c:95}
Yu, Q.; Yang, Y.; Liu, F.; Song, Y.-Z.; Xiang, T.; and Hospedales, T.~M.
  2016{\natexlab{b}}.
\newblock Sketch-a-Net: A Deep Neural Network that Beats Humans.
\newblock \emph{International Journal of Computer Vision}, 122: 411--425.

\bibitem[{Zhang et~al.(2016)Zhang, Liu, Zhang, Ren, Wang, and Cao}]{c:99}
Zhang, H.; Liu, S.; Zhang, C.; Ren, W.; Wang, R.; and Cao, X. 2016.
\newblock SketchNet: Sketch Classification with Web Images.
\newblock In \emph{Computer Vision and Pattern Recognition}.

\bibitem[{Zhang et~al.(2018)Zhang, Shen, Liu, Zhu, Yu, Shao, Shen, and
  Gool}]{c:83}
Zhang, J.; Shen, F.; Liu, L.; Zhu, F.; Yu, M.; Shao, L.; Shen, H.~T.; and Gool,
  L.~V. 2018.
\newblock Generative Domain-Migration Hashing for Sketch-to-Image Retrieval.
\newblock In \emph{ECCV}.

\end{thebibliography}

\bigskip

\end{document}